\documentclass[letterpaper]{article} 
\usepackage{aaai25}  
\usepackage{times}  
\usepackage{helvet}  
\usepackage{courier}  
\usepackage[hyphens]{url}  
\usepackage{graphicx} 
\usepackage{natbib}  
\usepackage{caption} 
\usepackage{algorithm}
\usepackage{algorithmic}
\usepackage{multirow}
\usepackage{booktabs}
\usepackage{amsmath}
\usepackage{amssymb}
\usepackage{mathrsfs}
\usepackage{caption}
\usepackage{subcaption}
\usepackage{tikz}
\usepackage{adjustbox}
\usepackage{newfloat}
\usepackage{listings}
\DeclareCaptionStyle{ruled}{labelfont=normalfont,labelsep=colon,strut=off} 
\floatstyle{ruled}
\newfloat{listing}{tb}{lst}{}
\floatname{listing}{Listing}
\title{OLMD: Orientation-aware Long-term Motion Decoupling for \\Continuous Sign Language Recognition}
\author{
Yiheng Yu,
Sheng Liu\footnote{Corresponding author},
Yuan Feng,
Min Xu,
Zhelun Jin,
Xuhua Yang
}
\affiliations{
    Zhejiang University of Technology\\
    \{yhyu, edliu, fengyuan, mxu2019, zljin, xhyang\}@zjut.edu.cn

}

\usepackage{bibentry}

\begin{document}

\maketitle

\begin{abstract}
The primary challenge in continuous sign language recognition (CSLR) mainly stems from the presence of multi-orientational and long-term motions. However, current research overlooks these crucial aspects, significantly impacting accuracy. To tackle these issues, we propose a novel CSLR framework: \textbf{O}rientation-aware \textbf{L}ong-term \textbf{M}otion \textbf{D}ecoupling (OLMD), which efficiently aggregates long-term motions and decouples multi-orientational signals into easily interpretable components. Specifically, our innovative Long-term Motion Aggregation (LMA) module filters out static redundancy while adaptively capturing abundant features of long-term motions.
 We further enhance orientation awareness by decoupling complex movements into horizontal and vertical components, allowing for motion purification in both orientations.
Additionally, two coupling mechanisms are proposed: stage and cross-stage coupling, which together enrich multi-scale features and improve the generalization capabilities of the model. Experimentally, OLMD shows SOTA performance on three large-scale datasets: PHOENIX14, PHOENIX14-T, and CSL-Daily. Notably, we improve the word error rate (WER) on PHOENIX14 by 
an absolute \textbf{1.6\%} compared to the previous SOTA.
\end{abstract}

%
\section{Introduction}
Continuous sign language features a distinctive grammatical structure and vocabulary \cite{cosign,Reagan_2007}, making it challenging for hearing individuals without prior knowledge to learn. Lately, various research has been devoted to CSLR to promote smooth communication between deaf and hearing people \cite{cosign,c2slr,cvtslr,vac,stmc}, significantly narrowing the gap between them.

\begin{figure}[t!]
    \begin{subfigure}[b]{0.23\textwidth}
   
    \includegraphics[width=\textwidth]{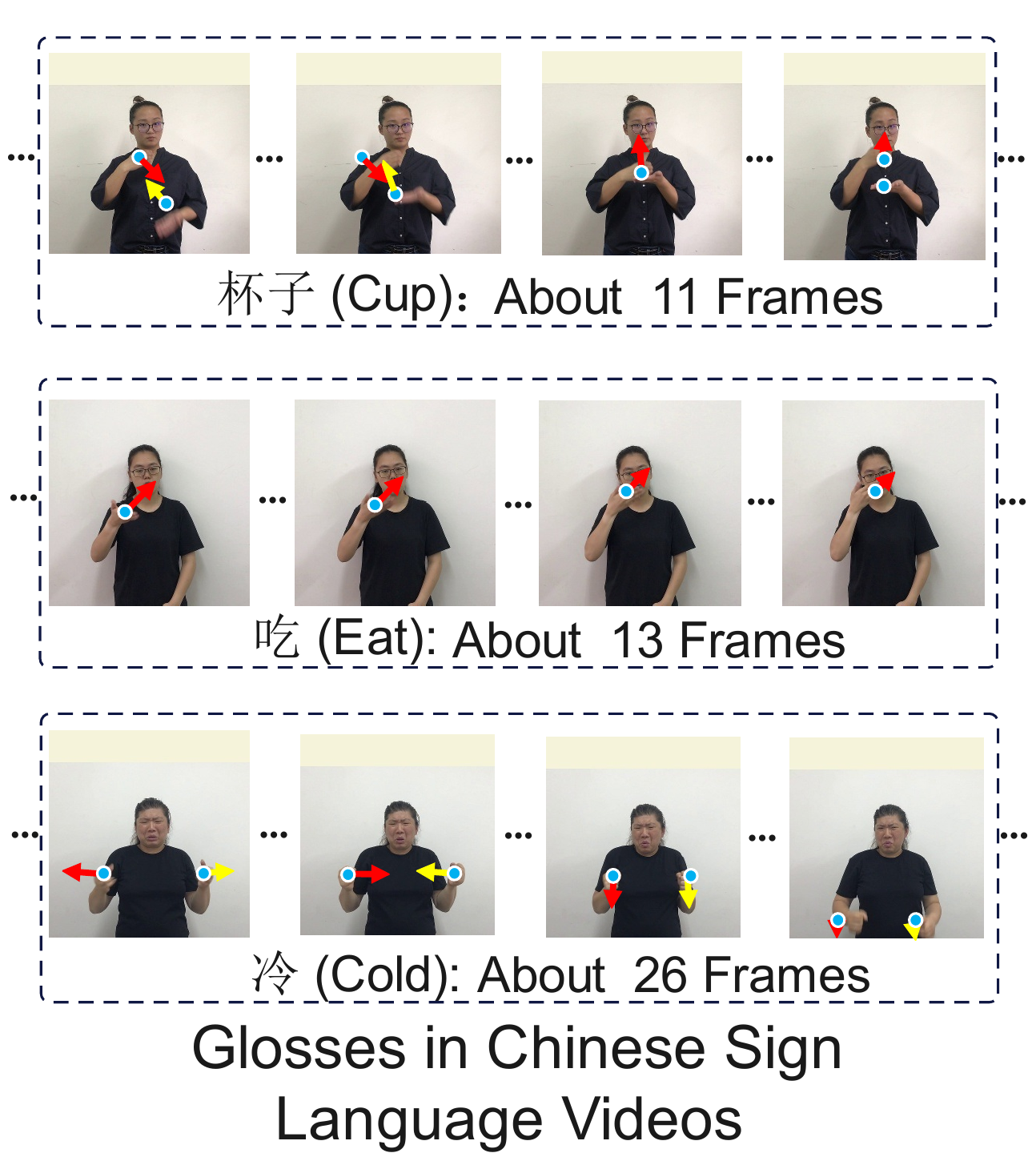} 
        \caption{}
        \label{fig:image1_1}
    \end{subfigure}%
    \begin{tikzpicture}[overlay, remember picture]
        \draw [dashed,line width=0.5mm] (0,4.8) -- ++(0,-4);
    \end{tikzpicture}%
    \begin{subfigure}[b]{0.25\textwidth}
        \includegraphics[width=\textwidth]{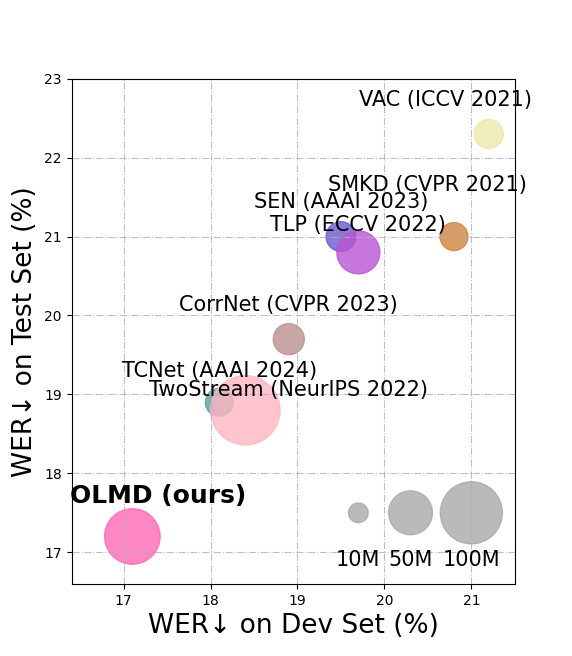}
        \caption{}
        \label{fig:image1_2}
    \end{subfigure}
    \caption{(a) We illustrate several common glosses with relevant video pieces from the CSL-Daily, highlighting the extensive presence of long-term and multi-orientational motions in sign language (yellow and red arrows indicate motion orientations), which traditional CSLR models struggle to manage. 
 (b) OLMD Performance Comparison on the PHOENIX14: surpassing  SOTA models by a large margin.}
  \vspace{-\baselineskip}
    \label{fig:combined1}
\end{figure}
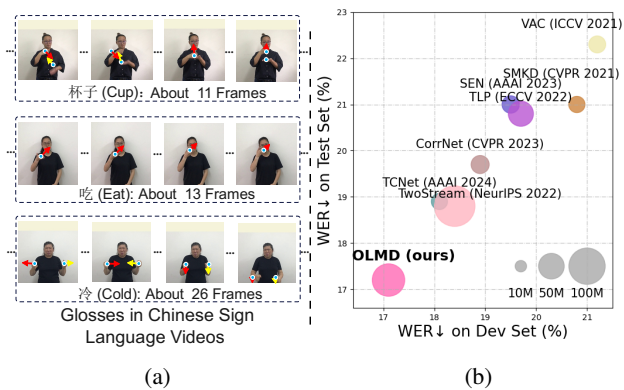



Recent CSLR research focuses on utilizing the motion inherent in sign language to improve recognition. For example, SEN \cite{sen} enhances the discriminative of keyframes by exploiting motion-induced differences, while CorrNet \cite{corrnet} calculates the correlation between adjacent frames to match areas with motion displacement. More recently, TCNet \cite{tcnet} captures motion trajectory through its optical-flow-inspired module. 
However, previous works ignore the long-term and multi-orientational nature of signs, which greatly affects the accuracy and robustness.


As shown in Fig. \ref{fig:image1_1}, complex multi-orientational long-term motions are prevalent in sign language. Typically, the movements forming a sign span 8-13 frames \cite{5206523}. However, current CSLR methods primarily analyze only 1-2 adjacent frames \cite{sen,corrnet,signgraph}, which is far from enough to capture a complete sign.

Additionally, much semantic information is embedded in orientations \cite{orientation2,orientation1}. To accurately decode this information, models should be orientation-aware. Unfortunately, conventional CSLR frameworks \cite{vac,smkd,corrnet} treat all orientations as a whole, resulting in confusion about orientational features and impacting accuracy.


To solve the above problems, we propose an \textbf{Orientation-aware Long-term Motion Decoupling (OLMD)}. Our approach includes a Long-term Motion Aggregation (LMA) designed to capture long-term motion information. The LMA enables OLMD to focus on dynamic regions and adaptively aggregate long-term motions, significantly improving its ability to capture entire signs.
  For multi-orientational motions, we introduce an orientational motion decoupling strategy. This approach decouples motion features into their horizontal and vertical components, allowing for a clearer analysis of each orientation’s contribution. By enhancing orientation awareness through motion purification, the method ultimately improves the recognition of multi-orientational signs.
   Additionally, stage and cross-stage coupling complete the decoupling-coupling process, where the latter enhances robustness by utilizing multi-scale features.

Extensive experiments show our proposed \textbf{OLMD} outperforms all previous models on the three widely-used datasets: PHOENIX14 \cite{Forster2015Continuous}, PHOENIX14-T \cite{8578910}, and CSL-Daily \cite{9578398}. Fig. \ref{fig:image1_2} highlights the excellent performance of OLMD on PHOENIX14.

Our main contributions are summarized as follows:
\begin{itemize}
    \item[$\bullet$] We innovatively design a Long-term Motion Aggregation (LMA) module, which effectively suppresses static redundancy while aggregating long-term motions in sign language.
    \item[$\bullet$]
We effectively decouple sign motions and enhance orientation awareness through orientation-aware motion purification, improving the model's understanding of complex signs.
    \item[$\bullet$] In extensive experiments, our OLMD demonstrates SOTA performance on three large datasets including PHOENIX14, PHOENIX14-T, and CSL-Daily.
\end{itemize}

\begin{figure}[t!] 
  \centering
  \includegraphics[width=0.45\textwidth]{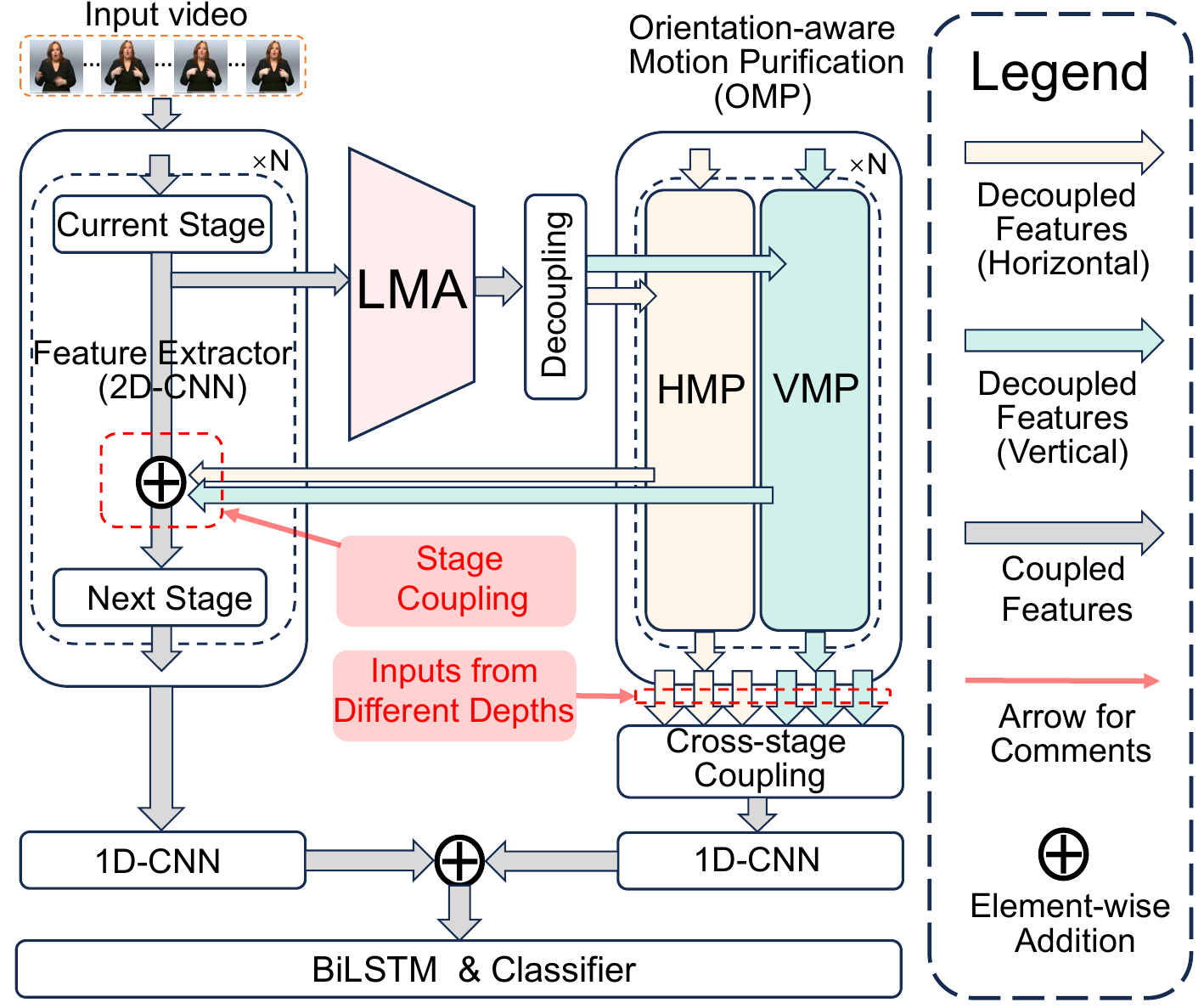}
  \caption{An overview of the proposed OLMD. After each stage of the Feature Extractor, frame-wise features first aggregate long-term motion information via LMA, which is then decoupled into horizontal and vertical components. HMP (Horizontal Motion Purification) and VMP (Vertical Motion Purification) are subsequently applied to enhance orientation-specific motion awareness. Stage and cross-stage coupling leverage enhanced features within and across stages, ensuring the integrity of decoupling-coupling while enriching the utilization of multi-scale features. Finally, two 1D-CNNs share architecture for downsampling and local temporal modeling, while the BiLSTM is used for global temporal modeling.}
  \label{fig:image2}
\end{figure}
\section{Related Work}
\label{sec:format}
\subsection{Continuous Sign Language Recognition}

Continuous sign language recognition (CSLR) \cite{cosign,cvtslr,vac,stmc,sen,corrnet}  aims to map untrimmed sign language videos to corresponding gloss\footnote{Gloss is the lexical unit of annotated continuous sign language.} sequences in a weakly-supervised way. Early CSLR works use hand-crafted features \cite{Cooper2009LearningSF,Yin2016IterativeRD} or HMM-based systems \cite{Koller2016DeepSH,Koller2017ReSignRE} to extract visual features and model long-time sequences. However, traditional handcrafted features often incur high computational costs and exhibit limited generalization performance. While HMMs struggle to capture nonlinear features and cannot be trained in an end-to-end manner. In recent years, deep learning methods have been extensively applied in CSLR, effectively addressing the limitations of traditional handcrafted features. Furthermore, the use of CTC loss \cite{ctc} facilitates end-to-end training of models. Overall, deep learning and CTC loss serve as the foundation for contemporary CSLR models.

Regarding inputs, present CSLR models can be broadly categorized into \textit{single-cue} and \textit{multi-cue}. \textit{Multi-cue} models, such as TwoStream \cite{twostream}, typically achieve high accuracy by incorporating rich information. However, these methods also introduce higher computational complexity and deployment costs. Considering the dual requirements of CSLR for real-time and accuracy, the current research strives to find a balance point based on enhancing the \textit{single-cue} models. For instance, VAC \cite{vac} and SMKD \cite{smkd} introduce alignment losses to optimize the synergy between local and global features of the model. CVT-SLR \cite{cvtslr} and CTCA \cite{ctca} utilize pre-trained linguistic knowledge to assist in the extraction of visual features. TLP \cite{tlp} refines the pooling layers, more intelligently extracting information across different levels of the temporal sequence. In general, these advances in \textit{single-cue} CSLR have laid the foundation for the introduction of \textbf{OLMD}.

\subsection{Motion Decoupling}
Motion decoupling has been effectively utilized in various fields to improve model performance and reduce noise. For example, \cite{decoupling1} employs a motion decoupling technique based on separating low-frequency and high-frequency signals, which reduces the noise interference in Video Motion Magnification (VMM). \cite{decoupling2} proposes a decoupling architecture based on distinct objects, aimed at estimating the motion of tissues and instruments under various constraints in medical imaging. \cite{decoupling3} proposes a decoupling architecture based on nonlinear Thin Plate Spline (TPS) transformation, capable of separating underlying motion features and guiding the synthesis of gesture videos. In this paper, we innovatively propose an orientational motion decoupling method. By decoupling complex motions into simple horizontal and vertical components, the model can enhance its orientation awareness through orientation-aware motion purification. As a result, our \textbf{OLMD} becomes better equipped to handle multi-orientational sign movements.

\section{Methods}
\subsection{Overview}
\begin{figure}[t!]
    \centering
    \begin{subfigure}[b]{0.235\textwidth}
        \includegraphics[width=\textwidth,height=3.8cm]{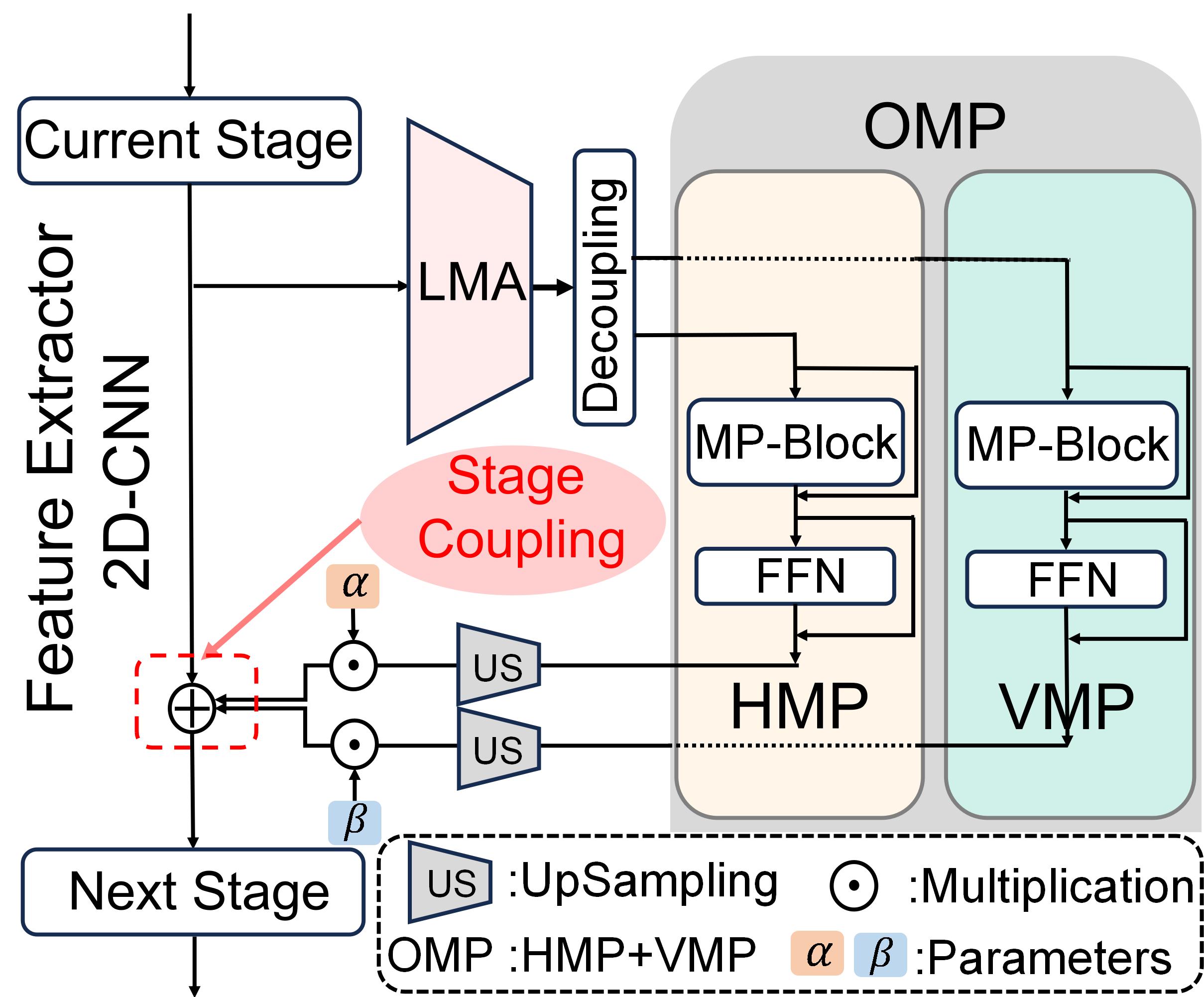}
        \caption{Non-cascaded design}
        \label{fig:image3_1}
    \end{subfigure}%
    \begin{subfigure}[b]{0.235\textwidth}
        \includegraphics[width=\textwidth,height=3.8cm]{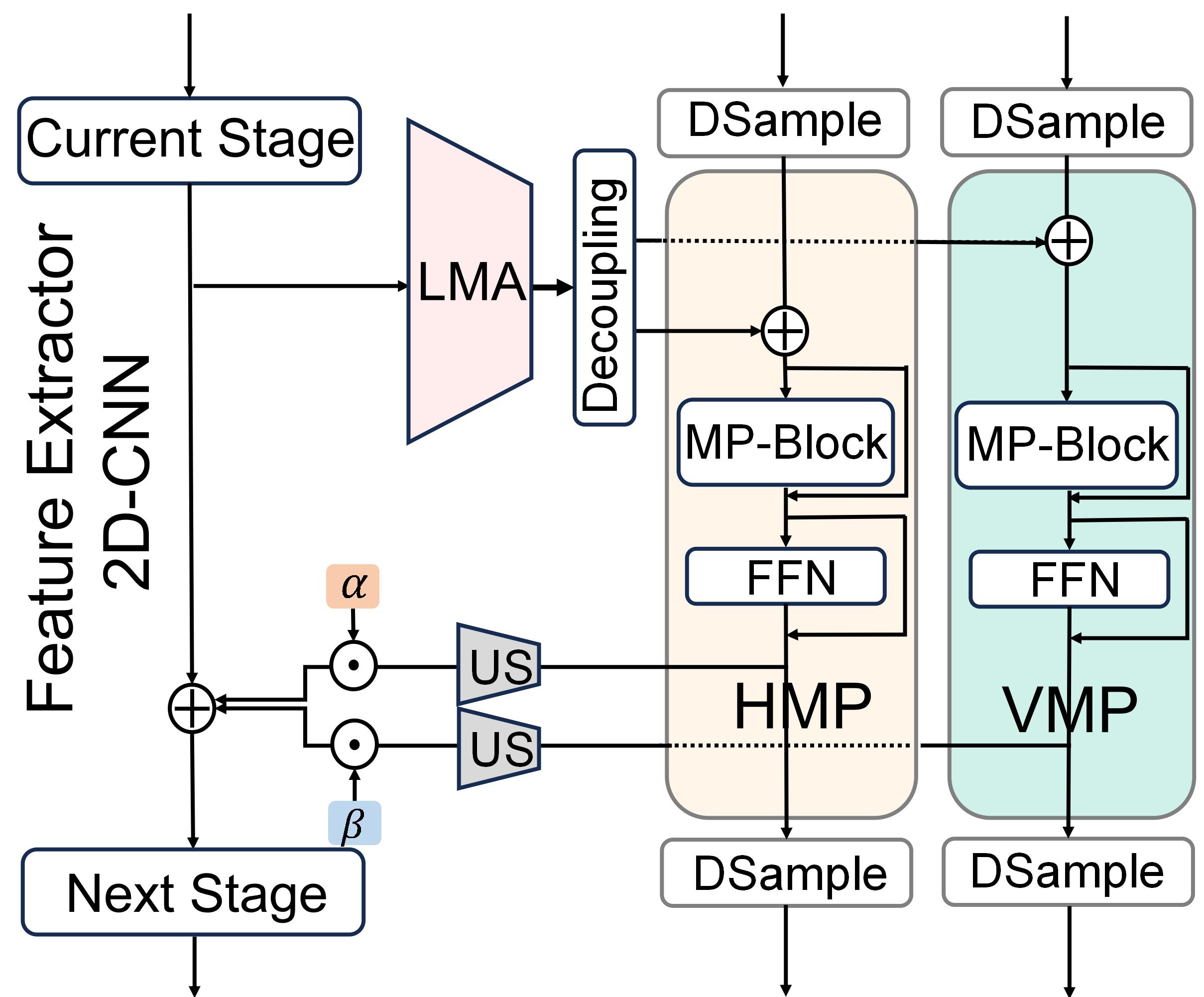}
        \caption{Cascaded design 
}
        \label{fig:image3_2}
    \end{subfigure}
    \caption{Details of the decoupling and stage-coupling are shown. (a) and (b) represent different designs of the OMP. }
     \vspace{-\baselineskip}
    \label{fig:combined2}
\end{figure}
The purpose of CSLR is to convert a video with $T$ frames $\boldsymbol{V}=\left\{v_{t}\right\}_{t=1}^{T}\in\mathcal{R}^{T\times H\times W\times3}$ into a series of glosses $\mathcal{G}=\{g_{i}\}_{i=1}^{N}$, where $N$ represents the length of the gloss sequence. Traditional methods employ a 2D-CNN to extract frame-wise features  $\boldsymbol{X} = \{ x_{t} \}_{t=1}^{T} \in \mathbb{R}^{T \times d}$. These features are then processed by a 1D-CNN and a BiLSTM for local and global temporal modeling. Finally, the outputs are classified while optimized by CTC loss.

Fig. \ref{fig:image2} shows \textbf{OLMD}, an innovative CSLR framework. After each stage of the Feature Extractor, we employ LMA to aggregate long-term motion information, followed by orientational motion decoupling. The decoupled features are then purified and enhanced in the VMP and HMP, respectively. Stage coupling integrates features within each stage before forwarding them to the next, while cross-stage coupling combines decoupled features from different depths.

\subsection{Decoupling-coupling Method}

Conventional models typically capture all motions as a whole \cite{vac,smkd}, which can result in feature confusion and reduce accuracy. In contrast, our method decouples motion features into their horizontal and vertical components, enhancing the model’s ability to discern complex patterns. Simply, the method consists of four parts: 1) motion decoupling, 2) orientation-aware motion purification, 3) stage coupling, and 4) cross-stage coupling. 

\textbf{Motion Decoupling.} As shown in Fig. \ref{fig:image3_1}, the frame-wise features $X\in\mathcal{R}^{C\times T\times H\times W}$
from the current stage first aggregate motion information to obtain $X_{agg}\in\mathcal{R}^{C\times T\times H\times W}$. The aggregated $X_{agg}$ encompasses multi-orientational motions.  To simplify these complex movements, we draw from physics-based methods by projecting motion features into a lower-dimensional space. This process decouples intricate motions into simpler, unidirectional components, making the dynamics clearer and more analyzable. Specifically, to isolate vertical motion, we exclude all horizontal components, and vice versa. The  process can be expressed as:
\begin{equation}\label{eq:1} 
\begin{split}
    X_{agg}=\mathrm{Agg}_{(k)}(X) ,\\
    X_{h},X_{v} = \mathcal{DP}(X_{agg}),
\end{split}
\end{equation}
where $\mathrm{Agg}_{(k)}(\cdot)$ denotes the aggregation of long-term motion (detailed in the next section), and $k$ represents the motion context length. $X_{h}\in\mathcal{R}^{C\times T\times W}, X_{v}\in\mathcal{R}^{C\times T\times H}$ denote the decoupled horizontal and vertical components, respectively. $\mathcal{DP}(\cdot)$ refers to a decoupling operation (e.g., Maxpool, Avgpool) identified in the ablation study.

\textbf{Orientation-aware Motion Purification (OMP).} OMP purifies the decoupled features, enhancing the model’s orientation awareness.
As shown in Fig. \ref{fig:image3_1}, OMP includes horizontal-aware motion purification (HMP) and vertical-aware motion purification (VMP) for purifying vertical and horizontal motions, respectively.
In HMP and VMP, the decoupled features $X_{h}$ and $X_{v}$ are treated as specialized 2D images. The MP-Block (2D CNNs, detailed in the supplementary materials) is employed to refine motion information in specific orientations, followed by a residual connection:
\begin{equation}\label{eq:2} 
\begin{split}
X_{h/v}=\mathrm{MP\text{-}Block}(X_{h/v})+X_{h/v},
\end{split}
\end{equation}
where $X_{h/v}\in\mathcal{R}^{C\times T\times (W / H)}$ denotes $X_{h}$ or $X_{v}$, encompassing only horizontal or vertical motions, respectively.

Subsequently, the FFN \cite{transformer} is employed to apply nonlinear transformations to channels, enhancing the representational capacity of OMP:
\begin{equation}\label{eq:3} 
\begin{split}
\mathrm{FFN}(X_{h/v})=\max(0,
\mathrm{Norm}(X_{h/v}W_1))W_2,
\end{split}
\end{equation}
where $W_1\in\mathcal{R}^{C_{hid}\times C_{in}\times 1\times1}$, $W_2\in\mathcal{R}^{C_{out}\times C_{hid}\times 1\times1}$ represent two 2D CNNs with a kernel size of 1, $\mathrm{Norm}(\cdot)$ represents layer  normalization.

Inspired by LGD \cite{lgd}, we also design the cascaded OMP. As shown in Fig. \ref{fig:image3_2}, decoupled features from the previous stage are downsampled and fused into the next, progressively enriching the information representation and increasing the network depth of OMP:
\begin{equation}\label{eq:4} 
\begin{split}
X_{h/v}=\mathcal{DS}(X_{h/v}^{pre})+X_{h/v},
\end{split}
\end{equation}
where $X_{h/v}^{pre}\in\mathcal{R}^{C\times T\times (W/H)}$ denotes the decoupled feature from the previous stage. $\mathcal{DS}(\cdot)$ denotes the downsampling method, which employs a 2D CNN with a stride of 2 in the last dimension.

 \textbf{Stage Coupling.}
The enhanced $X_{h}$, $X_{v}$ are coupled to update the features $X$ in the current stage of the Feature Extractor:
 \begin{equation}\label{eq:5} 
 X=X+\alpha\times \mathcal{US}(X_{h})+\beta\times \mathcal{US}(X_{v}),
\end{equation}
where $US(\cdot)$ represents the up-sampling operation (we repeat the features at each position). $\alpha$, $\beta$ are learnable parameters used to control the contribution of orientations. They are initialized to zero to preserve the original features.

\textbf{Cross-stage Coupling.}
Cross-stage coupling leverages multi-scale information by coupling features from different depths:
\begin{equation}\label{eq:6} 
X_{CS}=\sum_{i=1}^n(W_{i}(\mathcal{P}_{dim=W}(X_{h}^{i})+\mathcal{P}_{dim=H}(X_{v}^{i})),
\end{equation}
where $n$ depends on the number of decoupling iterations. $X_{h}^{i}$ and $X_{h}^{i}$ are derived from the $i$-th decoupling. $\mathcal{P}_{dim=H/W}(\cdot)$ represents the global average pooling operation on height or width dimension. $W_{i}$ denotes the $i$-th fully connected layer. $X_{CS}\in\mathcal{R}^{C\times T}$ denotes the features coupled through cross-stage coupling.

\begin{figure}[t!]
  \centering
  \includegraphics[width=0.4\textwidth]{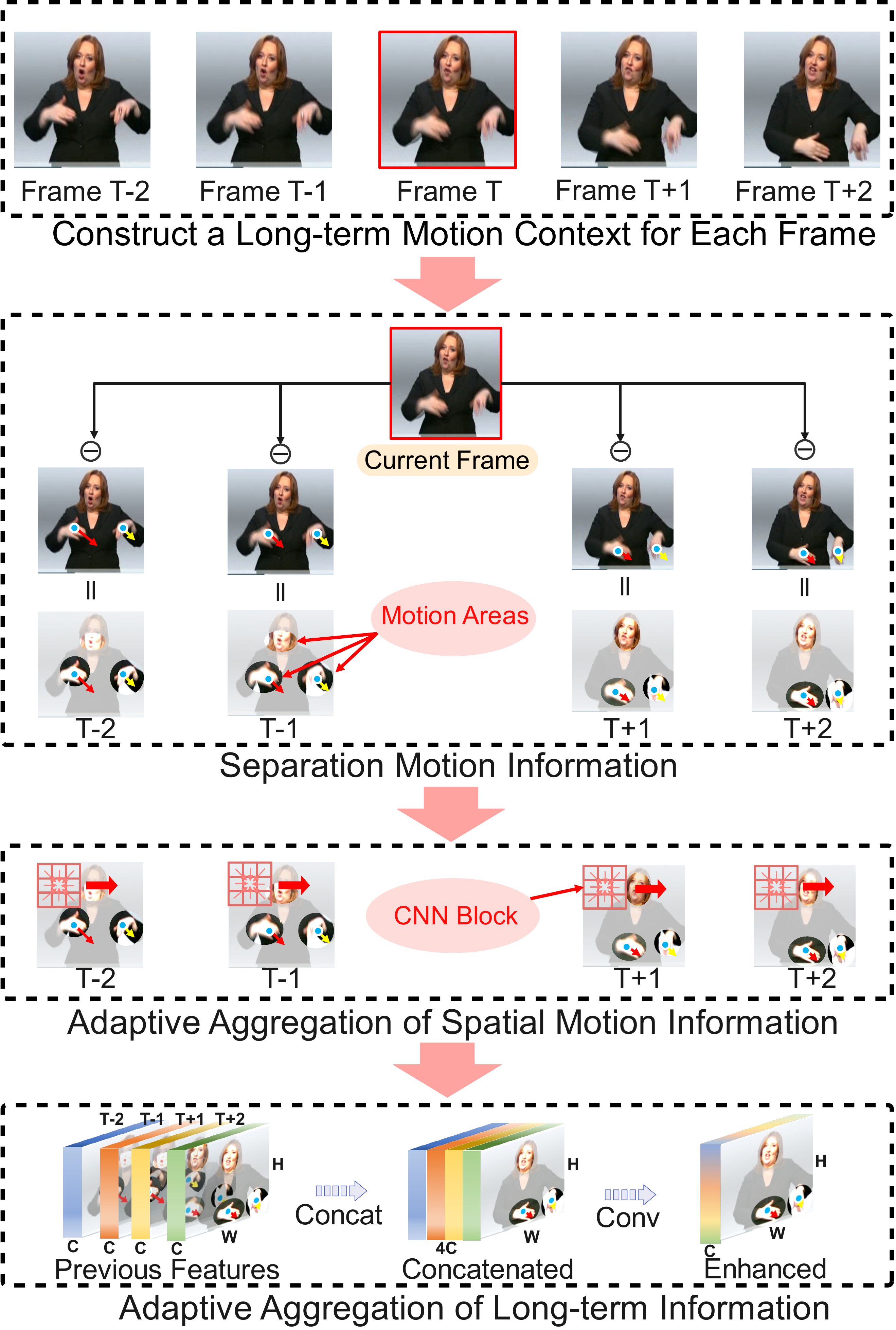}
\caption{The main idea of our Long-term Motion Aggregation (LMA) module, illustrated with a context length of 5. }
  \label{fig:image4}
\end{figure}

\subsection{Long-term Motion Aggregation}

Fig. \ref{fig:image1_1} illustrates the frequent long-term motions observed in sign language, while current works only focus on adjacent 1-2 frames \cite{sen,corrnet,signgraph}, which adversely impacts accuracy.
To tackle the problem, we propose the LMA module, as illustrated in Fig. \ref{fig:image4}. Each component is detailed below.

The $X = [x_1, x_2, \dots, x_T] \in \mathcal{R}^{C \times T \times H \times W}$ represents the sequence of features for $T$ frames, where each frame $x_i \in \mathcal{R}^{C \times H \times W}$. To reduce computational costs, the channel dimension is compressed to $C/r$ by a CNN.  Following this, we establish a long-term motion context for each frame through the following operation:

\begin{equation}\label{eq:7} 
 X_{(2n+1)}= \mathcal{S}_{(2n+1)}(X),
\end{equation}
where $\mathcal{S}_{2n+1}(\cdot)$  denotes the sliding window operation with a window size of $2n+1$, $X_{(2n+1)}\in\mathcal{R}^{(2n+1)\times C/r\times T\times H\times W}$ denotes the feature sequence with motion context $2n + 1$. 

We represent $X_{(2n+1)}$ unfolded along the motion context as $[\mathscr{C}_{(-n)},…,\mathscr{C}_{(0)},…,\mathscr{C}_{(n)}]$, where $\mathscr{C}_{(i)}\in\mathcal{R}^{C/r\times T\times H\times W}$ denotes the $i$-th frames offset from the current frames. The motion areas are subsequently separated through the following calculations:
\begin{equation}\label{eq:8} 
 \mathscr{C}_{(i)}^{'}(h,w)=\mathscr{C}_{(0)}(h,w)-\mathscr{C}_{(i)}(h,w), i \in [-n,n], i\neq0,
\end{equation}
$(h,w)$ denotes the spatial location,
if the feature at this location remains unchanged (e.g., background areas), the result will tend toward 0. This indicates that the model can focus on processing the key areas without being affected by redundancies.

To aggregate the discrete spatial motion regions and enhance the discriminative ability of the features, a shared CNN block is applied to sequentially process each frame:
\begin{equation}\label{eq:9} 
\mathscr{C}_{(i)}^{'}=\mathrm{ConvBlock}(\mathscr{C}_{(i)}^{'}), i \in [-n,n], i\neq0,
\end{equation}
where $\mathrm{ConvBlock}(\cdot)$ mainly consists of three CNNs, each with a kernel 
size of $1\times3\times3$.

Subsequently, $\{\mathscr{C}_{(-n)},...,\mathscr{C}_{(n)}\}$ are concatenated along the channel dimension, and a CNN is used to adaptively aggregate the long-term motion information while reducing the channels to $C/r$:
\begin{align}\label{eq:10}
X^{'}=\mathrm{Conv_{1\times1\times1}}\big(\text{Concat}\big(\big[\mathscr{C}_{(-n)},...,\mathscr{C}_{(n)} \big]\big)\big),
\end{align}
where $X^{'}\in\mathcal{R}^{C/r\times T\times H\times W}$.
Finally, we restore the channels back to $C$ with another CNN and apply a residual connection to retain the original information:
\begin{align}\label{11}
X^{out}=\mathrm{Conv_{1\times1\times1}}(X^{'})+X.
\end{align}



\begin{table*}[!t]
    \centering
   \setlength{\tabcolsep}{1mm}{

    \begin{tabular}{cllcccccc}
    \toprule
        \multirow{3}{*}{Methods} &   \multicolumn{2}{c}{\multirow{3}{*}{Cues}}&\multicolumn{4}{c}{PHOENIX14} & \multicolumn{2}{c}{PHOENIX14-T}\\
        \multicolumn{1}{c}{ }    &  & &\multicolumn{2}{c}{Dev(\%)} & \multicolumn{2}{c}{Test(\%)} & \multicolumn{1}{c}{Dev(\%)} & \multicolumn{1}{c}{Test(\%)}\\ \multicolumn{1}{c}{ }    &  & &del/ins & WER $\downarrow$ & del/ins & WER $\downarrow$ & WER $\downarrow$ & WER $\downarrow$\\
\midrule SFL \cite{sfl}&  \multicolumn{2}{c}{video}&7.9 / 6.5 & 26.2 & 7.5 / 6.3 & 26.8 & 25.1 & 26.1 \\
VAC~ \cite{vac}&  \multicolumn{2}{c}{video}&7.9 / 2.5 & 21.2 & 8.4 / 2.6 & 22.3 & 21.4 & 23.9 \\
SMKD~ \cite{smkd}&  \multicolumn{2}{c}{video}&6.8 / 2.5 & 20.8 & 6.3 / 2.3 & 21.0 & 20.8 & 22.4 \\
TLP~ \cite{tlp}&  \multicolumn{2}{c}{video}&6.3 / 2.8 & 19.7 & 6.1 / 2.9 & 20.8 & 19.4 & 21.2 \\
SEN~ \cite{sen}&  \multicolumn{2}{c}{video}&5.8 / 2.6 & 19.5 & 7.3 / 4.0 & 21.0 & 19.3 & 20.7 \\
 AdaBrowse+~ \cite{Hu2023AdaBrowseAV}&   \multicolumn{2}{c}{video}&6.0 / 2.5& 19.6&  5.9 / 2.6&  20.7& 19.5& 20.6\\
CorrNet~ \cite{corrnet}&  \multicolumn{2}{c}{video}&5.6 / 2.8 & 18.8 & 5.7 / 2.3 & 19.4 & 18.9 & 20.5 \\
CoSign~ \cite{cosign}&  \multicolumn{2}{c}{keypoints}&- & 19.7 & -  & 20.1 & 19.5 & 20.1 \\
SignGraph~ \cite{signgraph}& \multicolumn{2}{c}{video}&4.9 / 2.0 &18.2& 4.9 / 2.0 &19.1 &17.8 &19.1\\
 TCNet~ \cite{tcnet}&  \multicolumn{2}{c}{video}&5.5 / 2.4 & 18.1 & 5.4 / 2.0 & 18.9 & 18.3 &19.4 \\
\midrule CNN+LSTM+HMM*~ \cite{cnnlstmhmm}&  \multicolumn{2}{c}{video+mouth+hand}&-& 26.0& -& 26.0& 22.1&24.1\\
DNF*~ \cite{dnf}&  \multicolumn{2}{c}{video+optical flow}&7.3 / 3.3 & 23.1 & 6.7 / 3.3 & 22.9 & 22.7 & 24.0 \\
STMC*~ \cite{stmc}&  \multicolumn{2}{c}{ video+hand+face+pose}&7.7 / 3.4 & 21.1 & 7.4 / 2.6 & 20.7 & 19.6 & 21.0 \\
C$^{2}$SLR*~ \cite{c2slr}&  \multicolumn{2}{c}{video+keypoints}&-& 20.5 & -& 20.4 & 20.2 & 20.4 \\
 TwoStream-SLR*~ \cite{twostream}&  \multicolumn{2}{c}{video+keypoints}&-& 18.4& -& 18.8& 17.7&19.3\\
 \midrule  OLMD (ours)& \multicolumn{2}{c}{video}&\textbf{4.8} / 2.4&\textbf{17.1}&\textbf{4.7} / 2.3&\textbf{17.2}&\textbf{17.1}&\textbf{18.4}\\
 \bottomrule
\vspace{-\baselineskip}
 \end{tabular}
}
 
 \caption{Performance comparison with the latest methods on PHOENIX14 and PHOENIX14-T, * indicates the introduction of additional visual cues.
}
 \vspace{-\baselineskip}
\label{tab:1}
\end{table*}
\subsection{Loss Design}
The losses in OLMD are primarily divided into CTC losses and self-distillation losses.

\textbf{CTC losses.} CTC loss \cite{ctc} is the most widely used loss in CSLR tasks. It employs dynamic programming to address the alignment problem between outputs and labels, enabling end-to-end training. In OLMD, the total CTC loss is represented as:

\begin{equation}\label{eq:12} 
\mathcal{L}_{CTC}=\mathcal{L}_{CTC}^{local1}+\mathcal{L}_{CTC}^{local2}+\mathcal{L}_{CTC}^{global},
\end{equation}
where $\mathcal{L}_{CTC}^{local1}$ and $\mathcal{L}_{CTC}^{local2}$ represent supervision for the outputs of two 1D-CNNs, $\mathcal{L}_{CTC}^{global}$
represents supervision for the outputs of BiLSTM. Following SMKD \cite{smkd}, we share all classifiers before computing CTC losses.

\textbf{Self-distillation losses.} 
Inspired by VAC \cite{vac}, the outputs of the entire network are treated as the teacher, while the outputs of two 1D-CNNs are considered the students. The loss is computed using KL divergence:
\begin{equation}\label{eq:13} 
\mathcal{L}_{Distill}=\mathcal{L}_{KL}^{g2l 1}+\mathcal{L}_{ KL}^{g2l 2}.
\end{equation}

\textbf{Total loss.} The total loss is a combination of the CTC and self-distillation losses:
\begin{equation}\label{eq:14} 
\mathcal{L}=\gamma_{1}\mathcal{L}_{CTC}+\gamma_{2}\mathcal{L}_{Distill},
\end{equation}
where $\gamma_{1}$ and $\gamma_{2}$ are two hyperparameters used to balance contributions, we set them to 1 and 25, respectively.

\section{Experiments}
\label{experiments}
\subsection{Datasets and Evaluation}
In this paper, we mainly use three large CSLR datasets:

\textbf{PHOENIX14 \cite{Forster2015Continuous}} is a popular CSLR dataset from German TV weather reports with 1,295 words, recorded by 9 signers. It includes 5,672 training, 540 development (Dev), and 629 testing (Test) videos.

\textbf{PHOENIX14-T \cite{8578910}} is an expanded version of PHOENIX-2014, offering 7,096 training, 519 development (Dev), and 642 testing (Test) videos. It has a vocabulary of 1,115 sign language signs and 3,000 German words, serving sign recognition and translation tasks.

\textbf{CSL-Daily \cite{9578398}} is a large-scale Chinese sign language dataset filmed by 10 different signers, covering a variety of daily life themes with over 20,000 sentences. The dataset is split into 18,401 training samples, 1,077 development samples, and 1,176 test samples, featuring 2,000 sign language and 2,343 Chinese text vocabularies.

\textbf{Evaluation Metric.} We use word error rate (WER) as the performance metric for our model. It measures the minimum substitutions ($\#$sub), insertions ($\#$ins), and deletions ($\#$del) required to match a predicted sentence with the reference ($\#$ref):
\begin{equation}\label{eq:15} 
\mathrm{WER}=\frac{\#\mathrm{sub}+\#\mathrm{ins}+\#\mathrm{del}}{\#\mathrm{ref}},
\end{equation}
note that the \textbf{lower} WER, the \textbf{better} accuracy.

\subsection{Implementation Details}
This section primarily details the implementation of OLMD.

\textbf{Network details.} For Feature Extractor, we employ a ResNet34 \cite{resnet}, which is pretrained on ImageNet \cite{imagenet}.  For 1D-CNNs, we adopt SOTA configurations $\{K5, P2, K5, P2\}$, where $K5$ represents a convolution with a kernel size of 5, and $P2$ represents pooling with a kernel size of 2. We adopt the pooling method from TLP \cite{tlp}, setting the hidden layers of the BiLSTM to 1024. 
\textbf{Training and test strategy.} During training, we set the batch size to 2 and the initial learning rate to 0.001, reducing to 30\% at epochs 25 and 40. We default to using the Adam optimizer with a weight decay of 0.001, iterating for a total of 70 epochs.  All input frames are first resized to 256x256 and then randomly cropped to 224x224 during training, with a 50\% chance of horizontal flipping and a 20\% probability of temporal scale adjustment. For inference, we simply use a central crop of 224x224. Finally, all the training and testing are completed on 1 NVIDIA A6000 GPU.

\begin{table}[!htb]

\centering 
   \setlength{\tabcolsep}{1mm}
    \begin{tabular}{ccc}
    \toprule
    Methods &Dev(\%) & Test(\%) \\ 
    \midrule LS-HAN~ \cite{lshan} &39.0 & 39.4 \\
 SEN~ \cite{sen}&31.1&30.7\\
 AdaBrowse+~ \cite{Hu2023AdaBrowseAV}&31.2&30.7\\
    CorrNet~ \cite{corrnet} &30.6 & 30.1 \\
 CoSign \cite{cosign}& 28.1&27.2\\
    TCNet~ \cite{tcnet} &29.7 & 29.3 \\
    SignGraph~ \cite{signgraph} &27.3 & 26.4 \\
    \midrule
    TwoStream-SLR*~ \cite{twostream}&25.4 &25.3\\ 
    \midrule OLMD (ours)&\textbf{25.8}& \textbf{24.7}\\
    \bottomrule
    \end{tabular}
 
 \caption{Performance comparison with the latest methods on CSL-Daily, * indicates the introduction
of additional visual cues.}
\label{tab:2}
 \vspace{-\baselineskip}
\end{table}

\subsection{Comparison with SOTA Methods}
We verify the performance of OLMD on three large datasets.

\textbf{On PHOENIX14 and PHOENIX14-T.} Tab. \ref{tab:1} presents a comparison of OLMD with other SOTA models on PHOENIX14 and PHOENIX14-T datasets. All models are categorized into two groups: single-cue and multi-cue (denoted with *). The single-cue models only use RGB videos as input, while the multi-cue models use pre-extracted keypoints or heatmaps to aid recognition. 

Notably, OLMD outperforms all single-cue and even multi-cue models. For instance, compared to the single-cue SOTA  TCNet \cite{tcnet}, our WER on PHOENIX14 is reduced by 1.0\% (Dev) and 1.7\% (Test). On PHOENIX14-T, it's still reduced by 1.2\% (Dev) and 1.0\% (Test). 
The multi-cue SOTA Two-stream \cite{twostream} relies on auxiliary keypoint sequences and a large network structure to enhance recognition capabilities. However, OLMD surpasses it with single-cue input and a lightweight architecture. Remarkably, on the PHOENIX14, OLMD lowers the WER by 1.3\% (Dev) and 1.6\% (Test). The above results highlight the effectiveness of OLMD in capturing and leveraging complex movements in sign language, leading to a significant enhancement in performance.

\textbf{On CSL-Daily.} CSL-Daily is a challenging dataset with the largest vocabulary and the most sign language videos among all publicly available CSLR datasets. 

Tab. \ref{tab:2} shows our superior performance. Compared to the previous best single-cue method SignGraph \cite{signgraph}, OLMD exhibited a reduction in WER of 1.5\% (Dev) and 1.7\% (Test). Besides, we even show a WER decline of 0.6\% on the Test compared to the 
 best multi-cue method TwoStream. The above results indicate the advancement of OLMD and its adaptability to larger datasets.

\subsection{Ablation Studies}
We evaluate the effectiveness of each component via ablation studies on PHOENIX14.

\textbf{Ablations for the locations of decoupling.} In Tab. \ref{tab:3}, we ablate the decoupling positions after stages 2, 3, and 4. According to the results, decoupling at any position enhances performance, with more decoupling positions leading to more significant gains. Finally, OLMD applies decoupling after stages 2, 3, and 4, resulting in a 2.1\% improvement on Dev and 2.3\% on Test compared to the baseline.

\textbf{Ablations for LMA, HMP and VMP.}
Tab. \ref{tab:4} presents an ablation study of three key components: LMA, HMP, and VMP. LMA aggregates long-term information, while HMP and VMP purify vertical and horizontal motions, respectively. When features are decoupled and purified without LMA, performance improvements are trivial (0.3\% on Dev and 0.1\% on Test). However, integrating LMA with either HMP or VMP results in significant gains (2\% on Dev and 1.7\% on Test). The combination of all three components achieves the highest performance boost, particularly on the Test set, with an additional 0.7\% improvement.

\textbf{Ablations for LMA's motion context length \& parameters \& FLOPs \& inference time.}  
Tab. \ref{tab:5} presents the effects of varying motion contexts on model performance, parameters, FLOPs, and inference time. The results indicate a consistent decrease in WER with increasing context, while other metrics show varying increases. At a context length of 9, WER achieves its minimum, with parameters at 80.74M, FLOPs at 474.9G, and inference time at 0.047s. This demonstrates that the model remains both lightweight and real-time, leading us to select a context length of 9 for LMA.

\textbf{Ablations for two designs of OMP.} In Tab. \ref{tab:6}, we compare non-cascaded OMP with cascaded OMP. The results show that cascaded OMP performs better, particularly reducing WER by 0.9\% on the Test. This indicates that cascaded OMP progressively enhances features, increases the OMP's depth, and improves the model's generalization.

\textbf{Ablations for decoupling methods.}
In Tab. \ref{tab:7}, we evaluate various decoupling strategies: No-decoupling, MaxPool, AvgPool, and MaxPool+AvgPool. For the No-decoupling scenario, we adapted the 2D CNNs from OMP to 3D CNNs. The results demonstrate that all decoupling methods enhance performance relative to the No-decoupling approach, with AvgPool achieving the highest model performance. Consequently, we adopt AvgPool as the decoupling strategy for OLMD.

\textbf{Ablations for stage and cross-stage coupling.}
In Tab. \ref{tab:8}, both stage and cross-stage coupling methods outperform the uncoupled case. They yield the best results when combined, which is why OLMD incorporates them.

\begin{table}[t]
\setlength{\tabcolsep}{1mm}
    \centering{
    \begin{tabular}{ccccc}
        \hline
         Stage 2&Stage 3& Stage 4& Dev  (\%)& Test (\%)\\
        \hline
        -&-& -& 19.2& 19.5\\
 \checkmark & -& -&17.8 &18.4\\
          -&\checkmark & -& 17.5& 18.5\\
           -&- & \checkmark& 17.6&18.4\\
          \checkmark &\checkmark & -& 17.6& 17.9\\
 - & \checkmark & \checkmark & 17.3&17.7\\
           \checkmark &\checkmark & \checkmark & \textbf{17.1}& \textbf{17.2}\\
        \hline 
    \end{tabular} 
    }

\caption{Ablations for the locations of decoupling on the PHOENIX14. Stage n indicates decoupling after the n-th stage.}
\label{tab:3}
\vspace{-\baselineskip}
\end{table}

\begin{table}[!t]
\setlength{\tabcolsep}{1mm}
    \centering{
    \begin{tabular}{ccccc}
        \hline
         LMA&HMP& VMP& Dev  (\%)& Test  (\%)\\
        \hline
        -&-& -& 19.2& 19.5\\
 \checkmark & -& -& 18.3&19.3\\
          -&\checkmark & \checkmark & 18.9&19.4\\
          \checkmark &\checkmark & -& 17.2& 17.9\\
 \checkmark & -& \checkmark & 17.2&17.8\\
           \checkmark &\checkmark & \checkmark & \textbf{17.1}& \textbf{17.2}\\
        \hline 
    \end{tabular} 
    }

\caption{Ablations for LMA, HMP and VMP on the PHOENIX14.}
\label{tab:4}
\vspace{-\baselineskip}
\end{table}

\begin{table}[t]
\setlength{\tabcolsep}{1.3mm}{
\centering
\begin{tabular}{clccccc}
\hline
\textbf{Context} &  None&3& 5& 7& 9& 11\\ \hline
\textbf{Dev (\%)} &  18.9&18.0& 17.7& 17.5& \textbf{17.1}& 17.3\\
 \textbf{Test (\%)}& 19.4& 18.6& 17.8& 18.1& \textbf{17.2}&17.5\\
 \hline
\textbf{Params (M)}& 78.95& 80.08& 80.30& 80.52& 80.74&80.97\\
 \textbf{FLOPs (G)}& 363.4& 407.9& 430.3& 452.6& 474.9&497.2\\ 
\textbf{Time (S)}&  0.039&0.043& 0.045& 0.045& 0.047& 0.048\\ \hline
\end{tabular}}
\caption{ Ablations for context length of LMA on the PHOENIX14. We also record changes in parameters, FLOPs, and inference time (all tested on a 100-frame video).}
\vspace{-\baselineskip}
\label{tab:5}
\end{table}

\begin{table}[t]

    \centering
    \begin{tabular}{lcc}
        \hline
        Operation  & Dev (\%)& Test  (\%)\\
        \hline
 Non-cascaded  & 17.3&18.1\\
 Cascaded & \textbf{17.1}&\textbf{17.2}\\

    \hline
    \end{tabular}
    
    \caption{Ablations for two designs of OMP on the PHOENIX14.}
   \label{tab:6} 
   \vspace{-\baselineskip}
\end{table}

\begin{table}[t]

    \centering
    \begin{tabular}{lcc}
        \hline
        Operation & Dev (\%)& Test (\%)\\
        \hline
 Decoupling w/o & 18.2&18.5\\
 MaxPool& 18.0&18.4\\
 AvgPool& \textbf{17.1}&\textbf{17.2}\\
 MaxPool+AvgPool& 17.6&18.0\\
       \hline
    \end{tabular}
    
    \caption{Ablations for decoupling methods on the PHOENIX14.}
   \label{tab:7} 
   \vspace{-\baselineskip}
\end{table}

    

\begin{table}[!t]

    \centering
    \begin{tabular}{lcc}
        \hline
        Operation & Dev (\%)& Test (\%)\\
        \hline
 Coupling w/o &18.3 &18.5\\
 Stage& 17.9&18.3\\
 Cross-stage& 17.8&18.2\\
Stage + Cross-stage& \textbf{17.1}&\textbf{17.2}\\
       \hline
    \end{tabular}
    
    \caption{Ablations for stage and cross-stage coupling on the PHOENIX14.}
   \label{tab:8} 
   \vspace{-\baselineskip}
\end{table}

\section{Visualizations}
\begin{figure}[t!]
  \centering
  \includegraphics[width=0.4\textwidth]{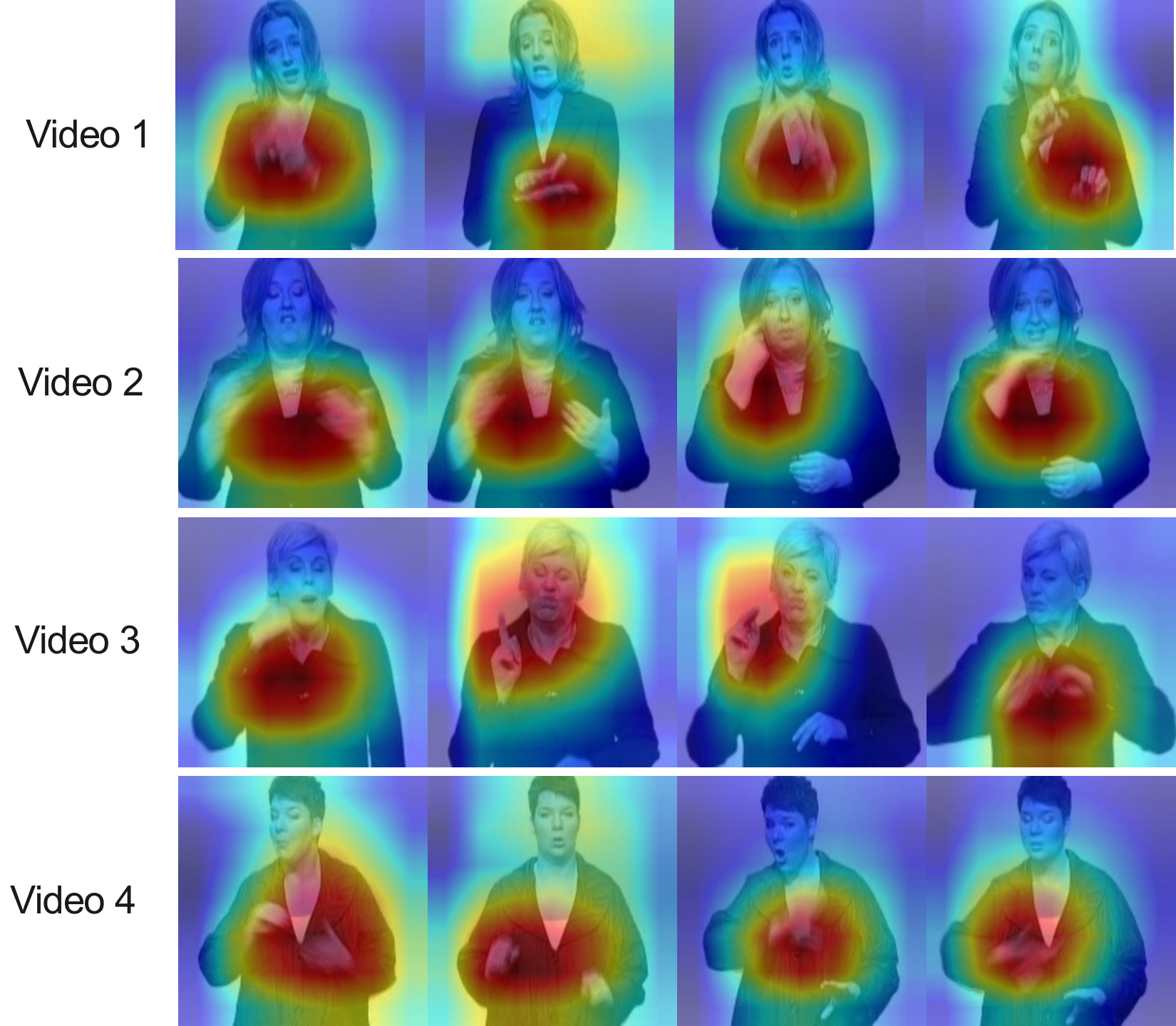}
\caption{Heatmap visualizations of the LMA module using Grad-CAM \cite{gradcam}. Obviously, LMA can effectively suppress static information (blue regions) and focus on motion areas (red and yellow regions, indicating moving hands and facial changes).}
  \label{fig:image5}
   \vspace{-\baselineskip}
\end{figure}

\textbf{Visualize for LMA.}  In Fig. \ref{fig:image5}, we present Grad-CAM \cite{gradcam} visualizations of the LMA applied to several sign language videos. The results show that the LMA  effectively captures the moving regions (especially the highly active hands) while suppressing attention to the background areas. This demonstrates that the LMA is well-designed to capture motion areas and suppress static regions.

\begin{figure}[t!]
  \centering
  \includegraphics[width=0.45\textwidth]{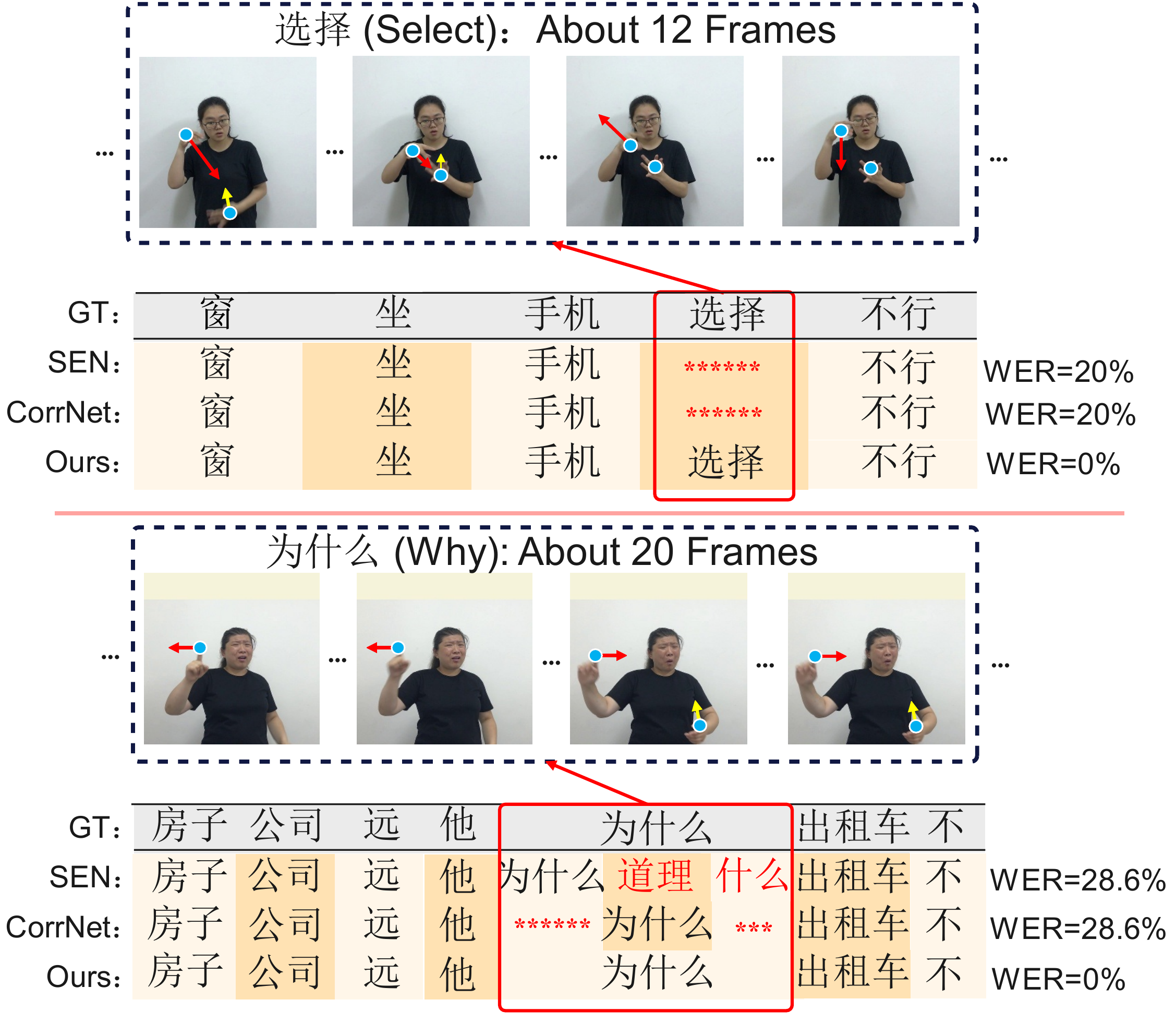}
\caption{Visualizations of two recognition examples on CSL-Daily (errors in red). \textbf{Above:} "Select" involves complex multi-orientational motions. SEN and CorrNet lacked orientation awareness, leading to incorrect recognition. OLMD enhances orientation awareness through decoupling and orientation-aware motion purification, enabling accurate recognition. \textbf{Below:} "Why" spans 20 frames. SEN and CorrNet failed to aggregate long-term motion, resulting in multiple glosses and increased errors. The LMA module enables OLMD to recognize this gloss accurately. }
  \label{fig:image6}
   \vspace{-\baselineskip}
\end{figure}

\textbf{Visualize for recognition results.} In Fig. \ref{fig:image6}, we showcase OLMD’s capability to recognize intricate signs. The “Select” involves rapid multi-orientational hand movements, posing a challenge for SEN and CorrNet to recognize it effectively. 
OLMD addresses this by decoupling motion features into horizontal and vertical components, enhancing orientation awareness through OMP. This approach significantly improves the model's ability to understand complex motions, enabling accurate recognition of "Select".
The “Why” lasts for 20 frames, which leads SEN and CorrNet to misidentify it as multiple glosses. However, OLMD correctly recognizes “Why” due to the LMA’s capability of aggregating long-term information.


\section{Conclusion}
This paper presents the OLMD, which markedly advances continuous sign language recognition (CSLR). The proposed LMA effectively aggregates long-term motions, enabling comprehensive sign capture. Furthermore, we enhance the model’s orientation awareness through decoupling and orientation-aware motion purification. Finally, stage and cross-stage coupling finalize the decoupling-coupling process, enhancing the robustness and accuracy. Experimental results on three large datasets demonstrate that OLMD significantly outperforms existing methods, establishing a robust new baseline for future research in CSLR.

\section{Acknowledgement}
This work is supported by the Natural Science Foundation of Zhejiang Province (LGG21F030011) and the National Key R\&D Program of China (No.2018YFB1305200).

\bibliography{aaai25}

\end{document}